# Understanding Stigmatizing Language Lexicons: A Comparative Analysis in Clinical Contexts

Yiliang Zhou, MS[1], Di Hu, MS[1], Tianchu Lyu, MBBS, MPH, PhD[1], Jasmine Dhillon, BS[1], Alexandra L. Beck, BS[1], Gelareh Sadigh, MD[1], Kai Zheng, PhD[1]
[1]University of California, Irvine, Irvine, CA, United States

**Abstract**

*Stigmatizing language results in healthcare inequities, yet there is no universally accepted or standardized lexicon defining which words, terms, or phrases constitute stigmatizing language in healthcare. We conducted a systematic search of the literature to identify existing stigmatizing language lexicons and then analyzed them comparatively to examine: 1) similarities and discrepancies between these lexicons, and 2) the distribution of positive, negative, or neutral terms based on an established sentiment dataset. Our search identified four lexicons. The analysis results revealed moderate semantic similarity among them, and that most stigmatizing terms are related to judgmental expressions by clinicians to describe perceived negative behaviors. Sentiment analysis showed a predominant proportion of negatively classified terms, though variations exist across lexicons. Our findings underscore the need for a standardized lexicon and highlight challenges in defining stigmatizing language in clinical texts.*

**Keywords.** natural language processing, health equity, stigma, stigmatizing language, electronic health record

**Introduction**

Disparities in healthcare outcomes among demographic groups in the United States have been well-documented[1]. Stigmatizing language by healthcare providers, defined as discrimination directed at a specific group of people, a place, or a nation[2], is one of the factors contributing to these disparities at both interpersonal and institutional levels[1,3]. Penn et al. found a more negative emotional tone in clinical notes among Black patients than those among White patients, suggesting the perpetuation of rooted systemic discrimination[4]. The same observation is seen among female patients (vs. male patients), patients with public insurance (vs. private), unmarried patients (vs. married), and patients with a higher Charlson Comorbidity Index[5–7]. Moreover, Park et al. identified negative emotional language in clinical notes as a form of stigmatizing language, further emphasizing the potential impact of biased documentation on patient care[8]. Goddu et al.'s study suggested that stigmatizing language in medical records could influence subsequent physicians-in-training, affecting their attitudes and medication-prescribing behaviors toward patients[9].

In recent years, research at the intersection of natural language processing (NLP) and clinical texts–including electronic health records (EHRs), free-text narratives, and other unstructured clinical documentation–has grown significantly[10–16]. The widespread adoption of EHRs has provided a rich and nuanced source of patient data that captures medical histories, physician assessments, and treatment plans[10,14,17–19], and NLP shows promise as an innovative approach to extracting and analyzing EHR data that conventional methods often miss[20]. This shift was largely accelerated by the Health Information Technology for Economic and Clinical Health (HITECH) Act, which aimed to promote the adoption and meaningful use of health technology[21]. Another point of acceleration is due to 21st Century Cures Act's implementation, which grants patients access to their EHRs through patient portals, allowing them to view physicians' notes[22]. As a result, researchers have increasingly leveraged these technological advancements to examine bias in clinical documentation.

Various techniques have been applied to analyze patterns of stigmatizing language within patient clinical text data. Himmelstein et al. found that 2.5% of the clinical notes in their sample contained stigmatizing language, with notes related to non-Hispanic Black patients showing a 0.67 percentage point greater likelihood of containing stigma compared to notes on non-Hispanic White patients[6]. Similarly, Weiner et al. identified stigmatizing language in 18.4% of analyzed clinical notes, notably finding that History and Physical notes were particularly prone to containing stigmatizing content[15]. Several studies have demonstrated the effectiveness of machine learning and large language models in accurately detecting and classifying stigmatizing language within EHRs[1,14,23], underscoring their potential to mitigate bias and improve health equity.

Lexicon has served as the backbone for previous studies to systematically detect and analyze biased language in clinical documentation, with researchers proposing their own stigmatizing language lexicons tailored to different tasks. However, discrepancies between lexicons may create confusion for future researchers in determining which terms to adopt in their analyses. Additionally, some lexicons–such as "clean" and "dirty" suggested by the National Institute on Drug Abuse (NIDA)–may not always clearly represent stigmatizing language depending on the context[2]. The literature is thus far silent on how existing stigmatizing language lexicons share similarities, which specific terms consistently appear across multiple lexicons, and the overall distribution of these terms in terms of positive, negative, or neutral sentiment. Therefore, this study aims to fill the gap by quantitatively reviewing existing stigmatizing language lexicons, focusing on three key aspects: (1) the extent to which these lexicons share similarities, (2) the specific terms that appear consistently across multiple lexicons, and (3) the distribution of these terms in terms of positive, negative, or neutral sentiment by comparing them with an established sentiment dataset.

**Methods**
*Search strategy*
We searched PubMed on February 28, 2025 using the following terms: ("stigmatizing language" OR "stigma" OR "biased language") AND ("electronic health record" OR "electronic health records" OR "electronic medical record" OR "electronic medical records" OR "EHR" OR "EMR" OR "text"[ti] OR "note"[ti] OR "narrative"[ti]) in the title and abstract, and included all work published since 2015 (inclusive). Three reviewers (Y.Z., J.D. and A.B.) collaboratively refined the search query, reached consensus, and then completed the search. This review focused on how prior studies defined and constructed stigmatizing language lexicons in clinical settings.

*Eligibility criteria*
We included articles meeting the following criteria: (1) studies focused on stigmatizing language within clinical documentation in a general healthcare context; (2) studies developing stigmatizing language lexicons; and (3) explicitly presented stigmatizing language lexicons either within the main manuscript or as supplemental materials. We excluded non-peer reviewed and non-English articles, as well as articles without full-text availability.

*Data extraction*
Stigmatizing language lexicons were extracted from each included study, specifically by retrieving all terms identified in the lexicons from the main manuscripts or their supplemental materials. Extracted terms within each lexicon were then organized into a structured lexicon dataset to facilitate subsequent analysis of similarities, sentiment, and frequency of stigmatizing language across studies.

*Semantic similarity analysis*
We utilized semantic similarity measures to assess the overlap of terms across stigmatizing language lexicons in the included studies. Initially, we employed WordNet to expand each term from the extracted lexicons into its corresponding sets of cognitive synonyms (synsets)[24]. Based on the evaluation from Slimani's study, we selected the Leacock-Chodorow (LCH) similarity measure, a path-based method that calculates semantic similarity by considering the length of the shortest path between two synsets within a hierarchical taxonomy[12,24]. Commonly, an LCH score above 2.5 suggests a strong conceptual relationship, while lower values indicate weaker associations[25]. To quantify the similarity between two lexicons, we computed the average pairwise semantic similarity between each term from the two lexicons using the following formula:

$$S_{avg}(L_1, L_2) = \frac{1}{|L_1| \times |L_2|} \sum_{W_1 \in L_1} \sum_{W_2 \in L_2} S_{LCH}(W_1, W_2)$$

Where $L_1$ and $L_2$ represent the two lexicons being compared, and $W_1$ and $W_2$ represent individual terms from lexicons $L_1$ and $L_2$, respectively.

To analyze word frequencies across the extracted lexicons, we first combined all the terms from the included stigmatizing language lexicons into a single master list and removed duplicates. For each term, we calculated how many individual lexicons contained that term. Then we applied a majority-vote threshold : if a term appeared in at least half (>=50%) of the lexicons, we defined it as high-frequency, indicating strong consensus and confirming it as stigmatizing language in clinical settings.

*Sentiment analysis*
To evaluate the sentiment of terms within the extracted stigmatizing lexicons, we utilized the WKWSCI Sentiment Lexicon, which contains a total of 29,718 words, including 3,121 positive words, 7,100 negative words, and approximately 19,500 neutral words[26]. For each extracted lexicon, we classified every included term according to its sentiment category (i.e., positive, negative, or neutral) as defined by the WKWSCI lexicon. We then calculated the proportion of the terms in each sentiment category for each stigmatizing lexicon, allowing us to quantify and compare the emotional or attitudinal composition across the lexicons.

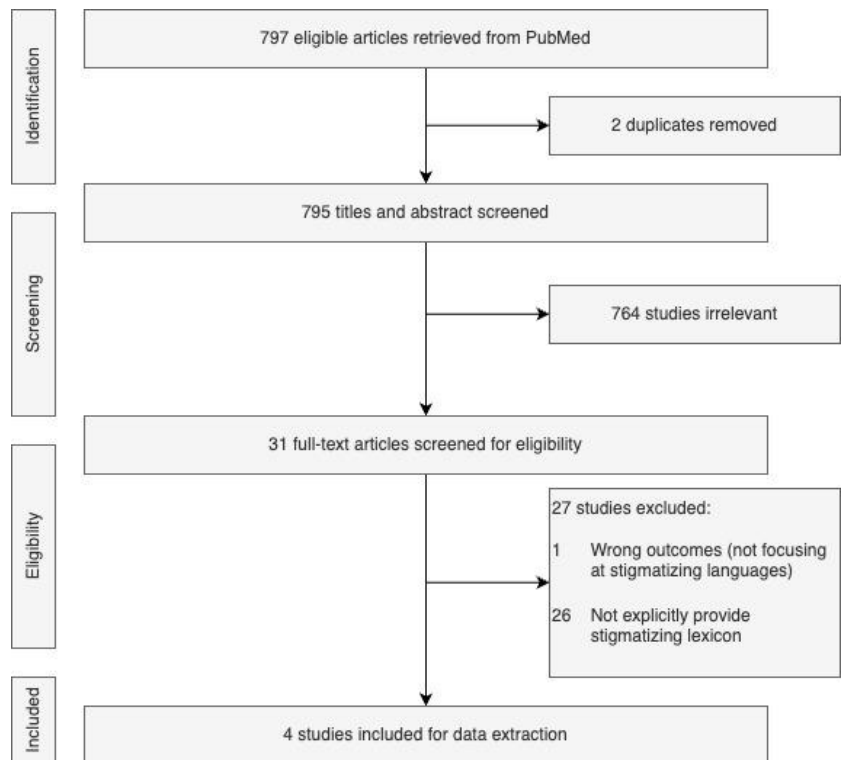

**Figure 1.** Search strategy and flow diagram for the stigmatizing language lexicon extraction process.

**Results**
The flow diagram (Figure 1) summarizes our study selection process. We initially identified 797 studies from PubMed published since 2015, of which two duplicate references were detected and removed using Covidence. After screening the titles and abstracts, we included 31 publications for a detailed full-text review for eligibility. Ultimately, four studies (Table 1) met our inclusion criteria and were selected for the main analysis, from which we extracted stigmatizing language lexicons for subsequent semantic similarity, word frequency, and sentiment analysis.

**Table 1.** Summary of the literatures with stigmatizing language lexicon.

| Study ID | Study Objective | Lexicon Size |
|---|---|---|
| Himmelstein 2022[6] | Examines stigmatizing language prevalence in hospital admission notes and patient and physician characteristics associated with stigmatizing language use. | 59 |
| Walker 2024[23] | Develops NLP methods to detect and classify stigmatizing language within EHRs. | 185 |
| Dickinson 2017[27] | Provides guidelines for stigmatizing language avoidance to healthcare professionals and others toward people with diabetes. | 53 |
| Harrigian 2023[1] | Characterizes stigmatizing language using domain-informed NLP techniques in EHRs. | 77 |

Among the final four studies selected, Walker et al. and Harrigan et al. developed lexicons specifically designed to detect and classify characteristics of stigmatizing language in EHRs using NLP techniques[1,23]. Himmelstein et al. utilized NLP along with stigmatizing language lexicon to examine the prevalence of stigmatizing language within hospital admission notes, as well as patient and physician characteristics associated with its usage[6]. Notably, Dickinson et al. offered guidelines for stigmatizing language avoidance to healthcare professionals and others when discussing diabetes in both spoken and written communication[27]. Although Dickinson et al. focused specifically on stigma towards people with diabetes, only 14.1% (9 out of 64) of the terms were explicitly diabetes related. Therefore, we determined that its lexicon is appropriate for inclusion and removed terms related to diabetes in subsequent analyses.

We also noted that the lexicon developed by Harrigian et al. includes several positive descriptors (e.g., "pleasant", "lovely") that, while not stigmatizing, may reduce the objectivity of the lexicon when used for subsequent comparative analyses. Therefore, we removed those positive descriptors (7 out of 84), as well as two positive descriptors in Dickinson et al. (2 out of 64). Table 1 presents the lexicon size after filtering.

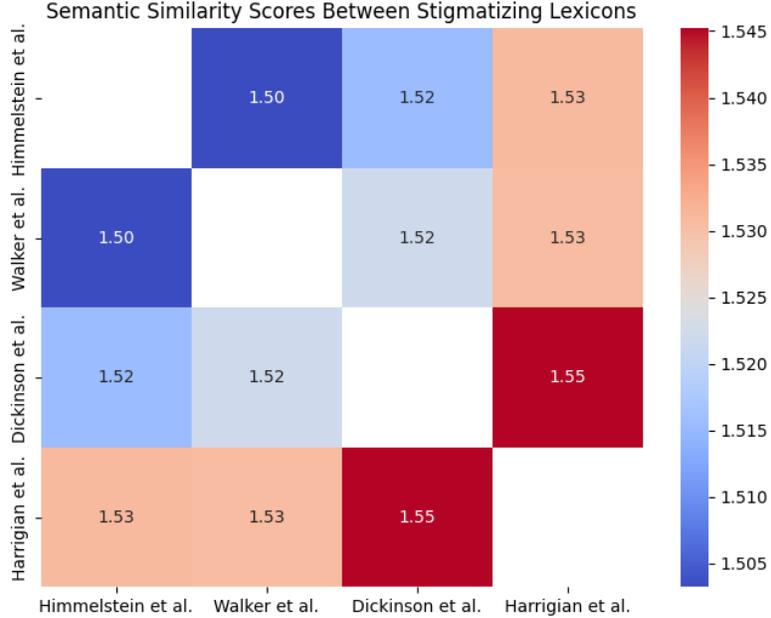

**Figure 2.** Heatmap of semantic similarity scores.
To further understand the overlap between stigmatizing language lexicons extracted, semantic similarity scores were measured for each pair of lexicons using the LCH similarity measure. Higher similarity scores indicated greater conceptual overlap and suggested shared semantic characteristics. The resulting pairwise similarity heatmap (Figure 2) illustrates to what extent each pair of lexicons aligns or differs in the representation of stigmatizing language. The average semantic similarity score among all four lexicons was only 1.52, indicating moderate semantic similarity between each pair. In other words, the lexicons shared both common themes and differences in defining stigmatizing language.

Word frequency analysis was conducted to identify commonly occurring stigmatizing terms across lexicons. After combining the extracted terms from all four included lexicons, a single master list consisting of 321 unique terms was generated. Among these, 45 terms appeared in at least two of the lexicons (Figure 3), indicating a moderate degree of overlap between the lexicons. Among the 45 high-frequency terms, 75.6% (34/45) were included in Harrigian et al., compared to 64.4% (29/45) om Himmelsteinet al., 53.3% (24/45) in Walker et al., and 44.4% (20/45) in Dickinson et al. Notably, three terms—"unwilling", "noncompliant", and "nonadherent"—were identified consistently across all four lexicons, highlighting their widespread recognition as stigmatizing language. Terms appearing frequently across lexicons were largely associated with patient behaviors judged negatively by clinicians, such as perceived non-adherence ("nonadherent", "noncompliant", "nonadherence"), refusal of care ("refused", "refusing", "in denial"), and argumentative behavior ("combatively", "defensive", "argumentative", "angry"). Additionally, terms like "difficult patient", "drug-seeking", and "lifestyle disease" represent stigmatizing judgments regarding patient character or responsibility for health conditions.

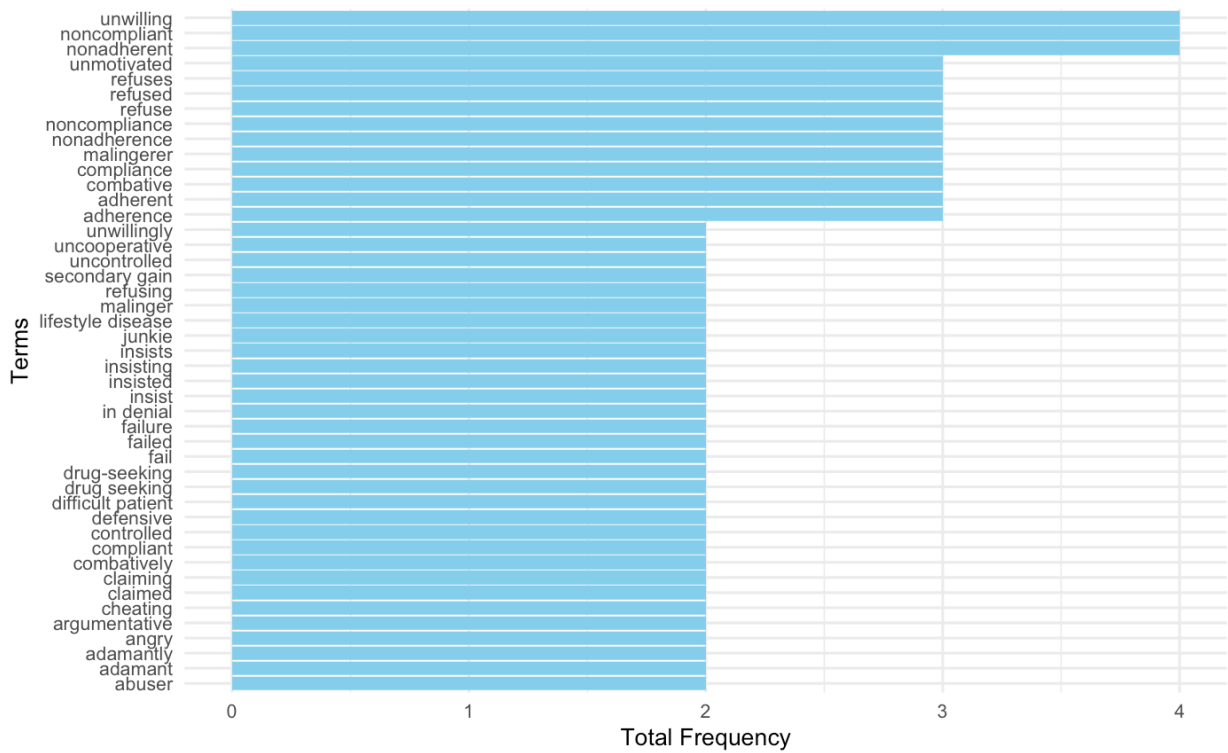

**Figure 3.** High-frequency stigmatizing language terms across included lexicons.

Sentiment analysis was performed using WKWSCI Sentiment Lexicon to classify terms from each stigmatizing lexicon into positive, negative, or neutral sentiment categories. Among the stigmatizing lexicons analyzed, 42.4% (25/59) of the terms from the lexicon by Himmelstein et al., 34.6% (64/185) from Walker et al., 39.6% (21/53) from Dickinson et al, and 46.8% (36/77) from Harrigian et al. matched terms in WKWSCI. The analysis (Table 2) showcased that the majority of matched terms were predominantly negative in sentiment (78.8% on average), while a smaller proportion were classified as positive (2.7%) or neutral (18.5%). Specifically, the proportion of negative sentiment terms varied notably across lexicons: approximately 93.8% (60/64) of matched terms from Walker et al. were negative, compared to 76% (19/25) from Himmelsteinet al., 69.4% (25/36) from Harrigian et al., and 52.3% (11/21) from Dickinson et al., highlighting differences in the extent of negativity across the stigmatizing lexicons. Among the 45 high-frequency terms, 77.3% (17/22) reflected negative sentiment, indicating that the different lexicons showed the greatest consensus on negatively connoted terms.

**Table 2.** Sentiment composition of stigmatizing language terms by lexicons.

| Study ID | Lexicon Size | Matched terms | Positive | Negative | Neutral |
|---|---|---|---|---|---|
| Himmelstein 2022[6] | 59 | 25 | 0 | 19 ("abuse", "addict", "cheat") | 6 ("adherence", "adherent", "habit") |
| Walker 2024[23] | 185 | 64 | 1 ("doggedly") | 60 ("doubtful", "dubious", "combative") | 3 ("insist", "speculative", "supposedly") |
| Dickinson 2017[27] | 53 | 21 | 1 ("compliant") | 11 ("bad", "suffer", "fail") | 9 ("prevent", "can", "compliance") |
| Harrigian 2023[1] | 77 | 36 | 2 ("cooperatively", "compliant") | 25 ("adamant", "refusal", "anger") | 9 ("claim", "adhere", "comply") |
| High-frequency terms | 45 | 22 | 1 ("compliant") | 17 ("junkie", "unwilling", "refuse") | 4 ("adherent", "compliance", "insist") |

**Discussion**

In this study, we quantitatively reviewed stigmatizing language lexicons from four studies to assess their similarities, overlap, and sentiment distribution. Our analysis revealed that the average LCH semantic similarity score among all four lexicons was only 1.52. Of the 321 unique terms identified across the four lexicons, only 45 terms appeared in at least two. Additionally, sentiment analysis showed that, on average, 78.8% of emotional terms were negative, 18.5% were neutral, and 2.7% were positive.

It was unsurprising that the average pairwise LCH semantic similarity remained moderate, indicating that while the lexicons share common conceptual themes, notable differences persist in their specific terms and definitions of stigmatizing language. This moderate overlap reflected variations in how researchers conceptualize stigmatization within clinical documentation. Such variability likely came from differences in their data sources used to curate lexicons and their study objectives, suggesting that nuanced variations in clinical texts can influence researchers' choices about which terms are identified as stigmatizing. For instance, Himmelstein et al. constructed their lexicon using free-text admission notes from a large academic medical center, tailoring it to identify stigmatizing language commonly found in admission documentation. In contrast, Harrigian et al. utilized data from a public dataset, MIMIC-III, and applied BioWordVec and GPT-3.5 to expand their initial lexicon, aiming to capture a more comprehensive list of identifiable stigmatizing patterns across EHR data. These differences highlighted the need for a universally accepted and standardized lexicon of stigmatizing language for future reference.

Within the master list of 321 unique terms compiled from the four lexicons from the selected studies, 45 were identified as high-frequency terms, appearing in at least half of the lexicons. Previous research showed consensus in labeling terms associated with clinician-judged negative behaviors as stigmatizing language[1,6,23,27]. The majority of these 45 high-frequency terms related to non-adherence, refusal of care, argumentative behavior, and judgments about patient identity and health conditions. These categories were consistent with several negative language categories introduced in Part et al.'s study[8], such as "Disapproval" and "Stereotyping." An interesting point was that, compared to stigmatizing language describing argumentative behavior or negative judgments about a patient's character, terms related to non-adherence and refusal of care were more commonly found across the four lexicons. While a patient's adherence to treatment is a crucial factor influencing health outcomes[28], the necessity of documenting non-adherence and exploring alternative terms remains an open discussion. Instead of labeling patients as non-adherent, a more objective approach would be to describe specific behaviors, such as the number of days medication was taken or the percentage of days it was taken as prescribed[27].

The proportion of negative sentiment terms varied across the four lexicons, ranging from 93.8% to 52.3%, further highlighting differences in how researchers curated their stigmatizing lexicons in terms of negativity. Despite the lexicon-level differences, those high-frequency terms revealed a broad consensus among researchers regarding the most negatively connoted language. While the predominance of negatively classified terms aligns with expectations for stigmatizing language, the presence of positive and neutral classifications underscored the complexity of identifying stigmatizing language based solely on sentiment. Positive terms, such as "pleasant" from Harrigian et al., often function for power/privilege language[3], demonstrating the challenge of differentiating them from stigmatizing language. Additionally, it reflected the fluidity of certain terms across different contexts. Some terms, such as "insist," can function as stigmatizing language—categorized by Park et al.[8] under "Disapproval"—or act as a neutral sentiment term depending on the scenarios.

There are several limitations to our study. First, we retrieved articles from only one electronic database, while relevant studies from other databases may have been overlooked. Second, our analysis included only four stigmatizing lexicons, which may limit the scope of comparison and fail to capture the full range of analysis of stigmatizing language used in clinical settings. However, these were the only lexicons that fit into our inclusion criteria and were available for analysis. Third, our study focused solely on individual terms, whereas stigmatizing language can also manifest at the contextual level, such as in full sentences, which we did not analyze. In future work, we will expand our literature search to additional databases, curate a comprehensive new lexicon of stigmatizing language, and integrate a human-in-the-loop verification process to ensure accuracy and contextual relevance. We will implement this lexicon within electronic health record systems to automatically flag stigmatizing terminology and leverage it as an educational tool for clinician training in health equity and culturally competent communication.

In general, our study reviewed lexicons from previous research and addressed the lack of a universally accepted and standardized lexicon of stigmatizing language, which is essential for consistency in research and health equity. Our findings provide a foundation for developing a more comprehensive lexicon and highlight key terms that consistently appear across existing lexicons. Additionally, we highlighted the challenges of identifying stigmatizing language solely through sentiment analysis and the fluidity of certain terms in different contexts. We call for further research to refine and standardize a universally accepted lexicon and to develop effective methods for detecting stigmatizing patterns in clinical texts. Leveraging advanced language models on more diverse datasets could uncover new linguistic characteristics at both the word and contextual levels, offering deeper insights into stigmatizing language that existing lexicons may not fully capture.

**Conclusion**

Overall, our study provides a quantitative review of stigmatizing lexicons within clinical documentation, highlighting both shared themes and notable differences across four existing lexicons. In the study, we explore whether there is semantic similarity among these lexicons, and identify high-frequency terms across multiple lexicons, which indicates a level of consensus in recognizing clinician-judged negative behaviors as stigmatizing language. However, the varying proportion of negative sentiment terms across lexicons, along with the presence of positive and neutral classifications, showcases the complexity of defining stigmatizing language purely based on sentiment. Our findings highlight the need for a more standardized and universally accepted lexicon of stigmatizing language, as well as methodological advancements for detecting stigmatizing patterns in clinical texts.

**References**

1. Harrigian K, Zirikly A, Chee B, Ahmad A, Links A, Saha S, et al. Characterization of Stigmatizing Language in Medical Records. In: Rogers A, Boyd-Graber J, Okazaki N, editors. Proceedings of the 61st Annual Meeting of the Association for Computational Linguistics (Volume 2: Short Papers) [Internet]. Toronto, Canada: Association for Computational Linguistics; 2023 [cited 2025 Mar 12]. p. 312–29. Available from: https://aclanthology.org/2023.acl-short.28/

2. Abuse NI on D. Words Matter: Preferred Language for Talking About Addiction | National Institute on Drug Abuse (NIDA) [Internet]. 2021 [cited 2025 Mar 12]. Available from: https://nida.nih.gov/research-topics/addiction-science/words-matter-preferred-language-talking-about-addiction

3. Barcelona V, Scharp D, Idnay BR, Moen H, Goffman D, Cato K, et al. A qualitative analysis of stigmatizing language in birth admission clinical notes. Nurs Inq. 2023;30(3):e12557.

4. Penn JA, Newman-Griffis D. Half the picture: Word frequencies reveal racial differences in clinical documentation, but not their causes. AMIA Summits Transl Sci Proc. 2022 May 23;2022:386–95.

5. Beach MC, Saha S, Park J, Taylor J, Drew P, Plank E, et al. Testimonial Injustice: Linguistic Bias in the Medical Records of Black Patients and Women. J Gen Intern Med. 2021 Jun;36(6):1708–14.

6. Himmelstein G, Bates D, Zhou L. Examination of Stigmatizing Language in the Electronic Health Record. JAMA Netw Open. 2022 Jan 27;5(1):e2144967.

7. Sun M, Oliwa T, Peek ME, Tung EL. Negative Patient Descriptors: Documenting Racial Bias In The Electronic Health Record. Health Aff Proj Hope. 2022 Feb;41(2):203–11.

8. Park J, Saha S, Chee B, Taylor J, Beach MC. Physician Use of Stigmatizing Language in Patient Medical Records. JAMA Netw Open. 2021 Jul 14;4(7):e2117052.

9. P. Goddu A, O'Conor KJ, Lanzkron S, Saheed MO, Saha S, Peek ME, et al. Do Words Matter? Stigmatizing Language and the Transmission of Bias in the Medical Record. J Gen Intern Med. 2018 May;33(5):685–91.

10. Bear Don't Walk OJ, Reyes Nieva H, Lee SSJ, Elhadad N. A scoping review of ethics considerations in clinical natural language processing. JAMIA Open. 2022 May 26;5(2):ooac039.

11. Barcelona V, Scharp D, Idnay BR, Moen H, Cato K, Topaz M. Identifying stigmatizing language in clinical documentation: A scoping review of emerging literature. PLOS ONE. 2024 Jun 28;19(6):e0303653.

12. Slimani T. Description and Evaluation of Semantic Similarity Measures Approaches. Int J Comput Appl. 2013 Oct 18;80(10):25–33.

13. Bilotta I, Tonidandel S, Liaw WR, King E, Carvajal DN, Taylor A, et al. Examining Linguistic Differences in Electronic Health Records for Diverse Patients With Diabetes: Natural Language Processing Analysis. JMIR Med Inform. 2024 May 23;12(1):e50428.

14. Barcelona V, Scharp D, Moen H, Davoudi A, Idnay BR, Cato K, et al. Using Natural Language Processing to Identify Stigmatizing Language in Labor and Birth Clinical Notes. Matern Child Health J. 2024 Mar 1;28(3):578–86.

15. Weiner SG, Lo YC, Carroll AD, Zhou L, Ngo A, Hathaway DB, et al. The Incidence and Disparities in Use of Stigmatizing Language in Clinical Notes for Patients with Substance Use Disorder. J Addict Med. 2023;17(4):424–30.

16. Bagheri A, Giachanou A, Mosteiro P, Verberne S. Natural Language Processing and Text Mining (Turning Unstructured Data into Structured). In: Asselbergs FW, Denaxas S, Oberski DL, Moore JH, editors. Clinical Applications of Artificial Intelligence in Real-World Data [Internet]. Cham: Springer International Publishing; 2023 [cited 2025 Mar 16]. p. 69–93. Available from: https://doi.org/10.1007/978-3-031-36678-9_5

17. Hirshman R, Hamilton S, Walker M, Ellis AR, Ivey N, Clifton D. Stigmatizing and affirming provider language in medical records on hospitalized patients with opioid use disorder. J Hosp Med. 2025;20(1):26–32.

18. Kyi K, Gilmore N, Kadambi S, Loh KP, Magnuson A. Stigmatizing language in caring for older adults with cancer: Common patterns of use and mechanisms to change the culture. J Geriatr Oncol. 2023 Nov 1;14(8):101593.

19. Yu E, Adams-Clark A, Riehm A, Franke C, Susukida R, Pinto M, et al. Perspectives on illness-related stigma and electronically sharing psychiatric health information by people with multiple sclerosis. J Affect Disord. 2021 Mar 1;282:840–5.

20. Mellia JA, Basta MN, Toyoda Y, Othman S, Elfanagely O, Morris MP, et al. Natural Language Processing in Surgery: A Systematic Review and Meta-analysis. Ann Surg. 2021 May;273(5):900.

21. Rights (OCR) O for C. HITECH Act Enforcement Interim Final Rule [Internet]. 2009 [cited 2025 Mar 12]. Available from: https://www.hhs.gov/hipaa/for-professionals/special-topics/hitech-act-enforcement-interim-final-rule/index.html

22. Arvisais-Anhalt S, Lau M, Lehmann CU, Holmgren AJ, Medford RJ, Ramirez CM, et al. The 21st Century Cures Act and Multiuser Electronic Health Record Access: Potential Pitfalls of Information Release. J Med Internet Res. 2022 Feb 17;24(2):e34085.

23. Walker A, Thorne A, Das S, Love J, Cooper HLF, Livingston M III, et al. CARE-SD: classifier-based analysis for recognizing provider stigmatizing and doubt marker labels in electronic health records: model development and validation. J Am Med Inform Assoc. 2025 Feb 1;32(2):365–74.

24. MIT Press [Internet]. [cited 2025 Mar 13]. WordNet. Available from: https://mitpress.mit.edu/9780262561167/wordnet/

25. Budanitsky A, Hirst G. Evaluating WordNet-based Measures of Lexical Semantic Relatedness. Comput Linguist. 2006;32(1):13–47.

26. Khoo CS, Johnkhan SB. Lexicon-based sentiment analysis: Comparative evaluation of six sentiment lexicons. J Inf Sci. 2018 Aug 1;44(4):491–511.

27. Dickinson JK, Guzman SJ, Maryniuk MD, O'Brian CA, Kadohiro JK, Jackson RA, et al. The Use of Language in Diabetes Care and Education. Diabetes Educ. 2017 Dec 1;43(6):551–64.

28. Horwitz RI, Horwitz SM. Adherence to treatment and health outcomes. Arch Intern Med. 1993 Aug 23;153(16):1863–8.